# Bio-Inspired Human Action Recognition using Hybrid Max-Product Neuro-Fuzzy Classifier and Quantum-Behaved PSO

Bardia Yousefi, Chu Kiong Loo
Department of Arti_cial Intelligence, Faculty of Computer Science and Information Technology,
University of Malaya, 50603 Kuala Lumpur, Malaysia
ckloo.um@um.edu.my

**Keywords:** biological inspired model, human action recognition, active basis model, optical flow, Synergetic neural network, quantum particle swarm optimization, Centroidal Voronoi Tessellations.

*Studies on computational neuroscience through functional magnetic resonance imaging (fMRI) and following biological inspired system stated that human action recognition in the brain of mammalian leads two distinct pathways in the model, which are specialized for analysis of motion (optic flow) and form information. Principally, we have defined a novel and robust form features applying active basis model as form extractor in form pathway in the biological inspired model. An unbalanced synergetic neural net-work classifies shapes and structures of human objects along with tuning its attention parameter by quantum particle swarm optimization (QPSO) via initiation of Centroidal Voronoi Tessellations. These tools utilized and justified as strong tools for following biological system model in form pathway. But the final decision has done by combination of ultimate outcomes of both pathways via fuzzy inference which increases novality of proposed model. Combination of these two brain pathways is done by considering each feature sets in Gaussian membership functions with fuzzy product inference method.*

*Two configurations have been proposed for form pathway: applying multi-prototype human action templates using two time synergetic neural network for obtaining uniform template regarding each actions, and second scenario that it uses abstracting human action in four key-frames. Experimental results showed promising accuracy performance on different datasets (KTH and Weizmann).*

*Povzetek:*

1. **INTRODUCTION**

Human brain is able to excellently recognize human object in different classes of action, recent methods are inspired by biological outcomes of computational neuroscience [32],[11]. In primary visual cortex (V1), procedure of images is more sensitive on bar-like structures. Responses of V1 are combined together by extrastriate visual areas and passed to inferotemporal cortex (IT) for tasks of recognition[10]. We follow the model of biological movement based on four assumptions, which are reliable by physiological and anatomical information [3]. The model splits to two corresponding pre-processing streams [1], [2], [12], [13], [61],[62,[63], [66] parallel to dorsal and ventral streams which are specified for analysis of flow and structure information, respectively. The model has used neural feature detector for extraction of optical flow and form features hierarchically considering size and style independency for both pathways, here we uses synergetic neural network in both feed-forward pathways for extraction of the structure and optical flow information. The corresponding results on the stationary human motion recognition revealing that discrimination can be accomplished through particularly small latencies, constructing an important role of top-down signals unlikely [1]. The proposed model expands an earlier model used for the stationary objects [13], [14], [10], [3], [8] recognition by adding and combining the information over time in dorsal and ventral pathway. Some visual physiologists have a regular belief regarding the proposed model [3]. It can be a good pertaining to quantity tool for organizing, summarizing and interpreting existent information. Initial structure design is based on data provided by neurophysiological. This developed structure implements quantitative estimation through computer simulations. Motion recognition and visual data has been involved in the model architecture. Proposed model has two separated pathways regarding form and motion information analysis. Information of two processing streams cooperates at few levels in mammalian brains [15], [16]. Mentioned coupling is able to ease the model integration for instance in STS level [17] and it develop the performance of recognition without varying the fundamental results.

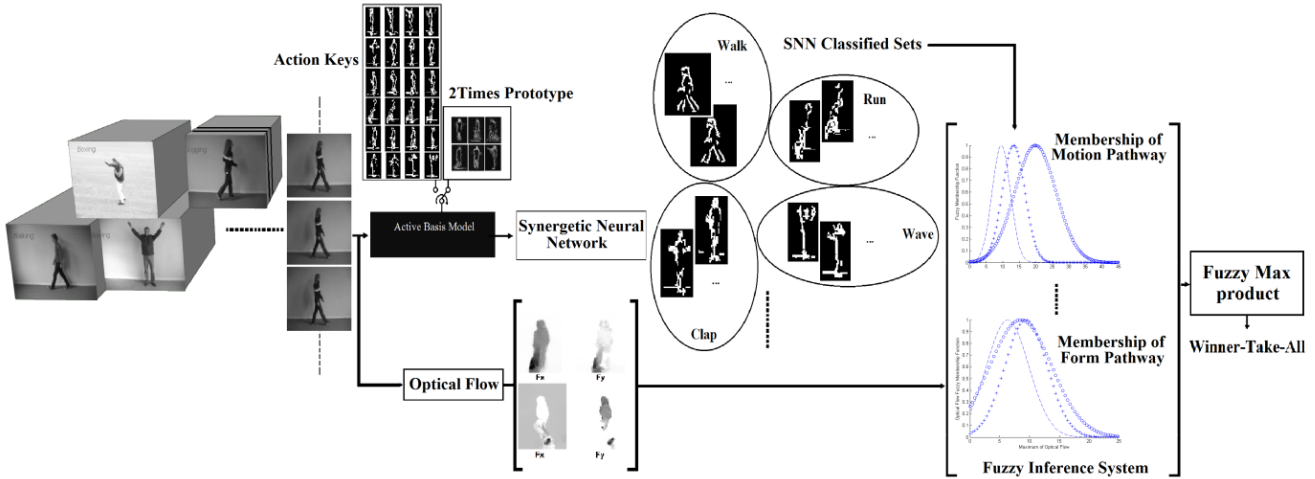

Figure 1: Gabor wavelet filter bank which has been used for active basis model is revealed.

**Form Pathway,** our proposed form pathway model follows an object recognition model [13] which is composed of form features detectors. It is capable to be reliable like data obtained from neurophysiological information concerning scale, position and sizes invariance which need further computational load along hierarchy. Modelling of cells in primary visual cortex (V1) in form pathway comprise detectors of local direction. Techniques having Gabor like filters for modelling the detectors has good constancy by simple cells [18].

Furthermore, neurons in monkey V1 range can influence the sizes of perceptive field in receptive fields [19]. Scale and location detectors are located in following level of this pathway that it finds information of local direction. There is an approximated independency for scales and spatial location inside receptive fields. Perhaps, complex-like cells in V1 area or in V2 and V4 are invariant regarding position varying responses (See [3]) and size independency is typically in area V4.

These two areas (V2, V4) are more selective for difficult form features e.g. junctions and corners whereas are not suitable for recognition of motion. To have an independent scale and position using mechanism of neuro-physiologically plausible model choosing detectors responses for different directions and receptive field scales and locations. Pooling achieved using maximum like operation (see [20]) some complex cells in cats visual cortex and areas V4 of macaques [21] reveal a maximum computing behaviour. Afterward, the snapshots detectors use for finding shapes of human body in area IT (inferotemporal cortex) of monkey where view-tuned neurons located and model of complex shapes tune[22]. Previous models used Gaussian Radial Basis functions for modelling and it adjusted within training which performed a key frame for training sequences. We develop neurobiological model [3],[7],[8] of processing of shape and motion in dorsal stream in visual cortex using active basis model [5] as computational mechanisms into the feed-forward aligned with motion pathway (optical flow).

**Motion Pathway,** In area MT and V1 there are some neurons for motion and direction selection in first level of motion pathway. There are many models for local motion estimation which are neurophysiologically plausible; we directly compute the response of motion-selective neurons by optical flow. Motion edges selectors in two opposite directions that it is found in MT, MSTd, MSTl and many parts of dorsal steams and probably in kinetic occipital area (KO) [3]. Proposed model, object specific motion position will be obtained by maximum pooling from motion position detector and considering motion selective edges which can be like MT [19] and MSTl [23] in macaque monkey. Motion pattern will be obtained considering membership functions related for every different action. Applying proposed approach is a simulation of both pathways in the primary visual cortex (V1) and projection of vertical stream in areas V2, V4, (see [3] and _g.1).

## 2. RELATED WORK

Human action recognition tasks generally categorize as two separated classes. First class prefers to track the part of image which is object (human) exists [24]. Mentioned groups of techniques might not useful in less articulated objects. However, they are considered as successful approaches. The other popular class is addressed on low resolution videos, high locally resolution images [6].or using spatiotemporal features [?]. As it has previously discussed regarding neurobiological inspired model for analysis of movement in dorsal stream visual cortex and psychological and physiological information; our proposed approach categorized as second group of methods. Previous method [3] has constant translation lack and a limited hand-crafted features dictionary in intermediate periods [25]. Jhuang et al. (2007) [7] and Schindler et al. [8] present successful biological inspired method for human action recognition.

Main contributions, In our proposed approach, major contribution is improving the neurobiological model which combination of two pathways is better done. Applying active basis model which makes form pathway more robust and developing the model applying fuzzy inference for aggregation of two pathways. For neuroscience model [3] into the real world by computer vision algorithm, two important techniques have been altered increasing performance in form pathway and developing the combination in two pathways. Besides, quantum particle swarm optimization for synergetic neural network represent plausible neurophysiological model.

## 3. MODEL OVERVIEW

The proposed system addresses a biological inspired system like [3], [9] and based on [3] which input is images obtained from video sequences.

### 3.1 Form features using Active basis Model

Active basis model [5] applying Gabor wavelets (for elements dictionary) offers deformable biological template. Shared sketch algorithm (SSA) followed through AdaBoost. In every iteration, SSA following matching pursuit chooses an element of wavelet. It checks the objects number in different orientation, location, and scale.

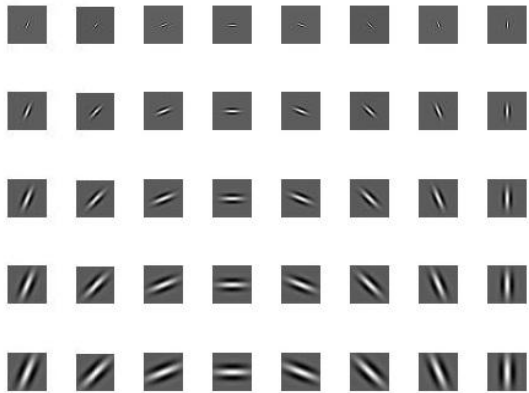

Figure 2: Gabor wavelet filter bank which has been used for active basis model is revealed.

Selecting the small number of elements from the dictionary for every images (Sparse coding), therefore there can be representation of image using linear combination of mentioned elements by considering U as a minor residual.

$$I = \sum_{i=1}^{n} c_i \beta_i + \varepsilon \quad (1)$$

Where $\beta = (\beta_i, i = 1,...,n)$ is set of Gabor Wavelet elements and components of sin and cosine, $c_i = \langle I, \beta_i \rangle$ and $\varepsilon$ is unsolved image coefficient [5]. By using wavelet sparse coding large number of pixels reduces to small number of wavelet element. Sparse coding can train natural patches of image to a Gabor like wavelet elements dictionary which carries the simple cells in V1 properties [11, 5]. The extraction of local shapes will be separately done for every frame and like [8] responses of filter orientation and density of each pixels computes. Also, the active basis model [5] uses the Gabor filter bank but in different form.

A Gabor wavelets dictionary, comprising n directions and m scales is in the form of, $GW_j(\theta, \omega)$, $j = 1,...,m \times n$.

Where, $\theta \in \left\{\frac{k\pi}{n}, k = 0,...,n-1\right\}$ and $\omega = \left\{\frac{\sqrt{2}}{i}, i = 1,...,m\right\}$.

Gabor wavelet features signifies the object form as small variance in size and location and posture. Though overall shape structure, it considers to be maintained throughout the process of recognition. Response (convolution) to each element offers form information with $\theta$ and $\omega$.

$$B = <GW, I> = \sum\sum GW(x_0 - x, y_0 - y : \omega_0, \theta_0) I(x, y). \quad (2)$$

Let $GW_j$ is a $[x_g, y_g]$ and $I$ is a $[x_i, y_i]$ matrixes, response of $I$ to $GW$ is a $[x_i+x_g, y_i+y_g]$. Therefore, previous convolution both matrices must be padded through sufficient zeros. Consequence of convolution can be eliminated via cropping the result. Additional technique would be to shift back the center of the frequencies (zero frequency) to center of the image though it might reason for loosing data.

Obtained training image set $\{I^m, m = 1, ..., M\}$, the joint sketch algorithm consecutively chooses $B_i$. The fundamental opinion is to find $B_i$ so that its edge segments obtain from $I_m$ become maximum [5]. Afterward, it is necessary to compute $[I^m.\beta] = \psi(|\langle I^m.\beta\rangle|^2)$ for different $i$ where $\beta \in$ Dictionary and $\psi$ represents sigmoid, whitening, and thresholding transformations and. Then for maximizing $[I^m.\beta]$ for all possible $\beta$ will be computed. Let $\beta = (\beta_i, i = 1, ..., n)$ is the template, for every training image $I_m$ scoring will be based on:

$$M(I^m, \beta) = \sum_{i=1}^{n} \delta_i |I^m, \beta_i| - \log \Phi(\lambda \delta_i)). \quad (3)$$

M is the match scoring function and $\sigma_i$ obtained from $\sum_{m=1}^{M}[I^m, \beta_i]$ regarding steps selection and $\Phi()$ is nonlinear function. The logarithmic likelihood relation of exponential model attains from the score of template matching. Vectors of the weight calculate by maximum likelihood technique and are revealed by $\Delta = (\delta_i, i = 1, ..., n)$ [5].

$$MAX(x,y) = \max_{(x,y)\in D} M(I^m, \beta). \quad (4)$$

MAX(x,y) calculates the maximum matching score obtained previously. D represents the lattice of I. Here, there is no summation because of updating the size based on training system on frame (t-1). Moreover, the method tracks the object applying motion feature for getting displacement of moving object.

**3.2 Motion features.**
For having the features regarding the motion of subject, the layer-wised optical flow estimation has been done. A mask which reveals the each layer visibility is the main different between estimation of traditional and layer-wised optical flow. The mask shape is able to perform fractal and arbitrary and only matching applies for the pixels which fall inside the mask (see [4]).
We use the layer-wised optical flow method in [4] which has baseline optical flow algorithm of [26, 27, 28]. As an overview, M1 and M2 are visible masks for two frames

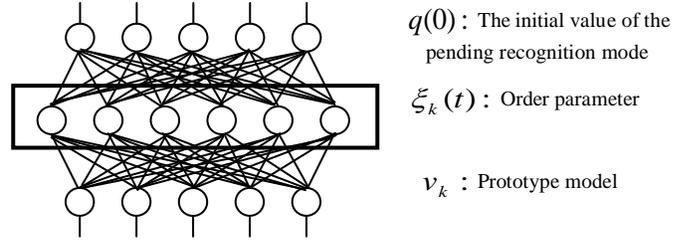

$q(0)$ : The initial value of the pending recognition mode

$\xi_k(t)$ : Order parameter

$v_k$ : Prototype model

Figure 3: The synergistic pattern recognition

I1(t) and I2(t-1), the field of flow from I1 to I2 and I2 to I1 are represent by(u1,v1), (u2,v2). Following terms will be considered for layer-wise optical flow estimation. Objective function consists of summing three parts, visible layer masks match to these two images using Gaussian filter which called data term matching $E_\gamma^{(i)}$, symmetric $E_\delta^{(i)}$, and smoothness $E_\mu^{(i)}$.

$$E(u_1, v_1, u_2, v_2) = \sum_{i=1}^{2} E_\gamma^{(i)} + \rho E_\delta^{(i)} + \xi E_\mu^{(i)}. \quad (5)$$

After optimization of objective function and using outer and inner fixed-point iterations, image warping and coarse to fine search, we attain flow for both bidirectional. Compressed optic flow for all the frames are calculated by straight matching of template to the earlier frame by applying the summation of absolute difference (L1-norm).

Though optic flow is particularly noisy, no smoothing techniques have been done on it as: the field of flow will be blurred in gaps and specially the places that information of motion is significant [7]. To obtain the proper response of the optical flow regarding its application in the proposed model, optical flow will be applied for adjust the active basis model and makes it more efficient.

To achieve a representation reliable through the form pathway, the optic flow estimates the velocity and flow direction. The response of the filter based on local matching of velocity and direction will be maximal as these two parameters are continuously changing.

**3.3 Synergetic neural network classifier**
Analyzing the human brain cognitive processes [45-48], particularly the visual analysis, we apprehend that the brain persistently involved in a big amount of the perception re-processing, subconscious mind, filtering, decomposition and synthesis. The brain of the human is a cooperative system, in some cases, cognitive processes can be supposed to depend on the self-organizing pattern formation. Based on this knowledge, Haken presents synergetic neural network as one the pattern recognition process which performs in brain of the human. A joint method to association of trained samples is values of feature averaging (see [47]). He revealed a collaborative pattern recognition of a top-down thinking: pattern recognition process can be comprehended like a specific order parameter competition process for recognition mode q can construct a dynamic process, so q after middle state q(t) into a prototype pattern vk.

Though it is not enough flexible for direction changing. Therefore the boundaries of these templates are not clear. Applying learning object in the same view is a technique for dealing with inflexibility which will limit the task of classification. Algorithm of melting is introduced by [29] for objects combination in diverse pose. Assume a trained object sample $I'_i$ contains of n pixel values. By reshaping $I'_i$ to $v_i$ which is a column vector matrix and normalization will have:

$$\sum_{j=1}^{n} v_{ij} = 0, \quad \sum_{j=1}^{n} v^2_{ij} = 1$$

Where q is the input mode q0 is the initial values of the state vector for attention parameters, which will discuss later, Connected prototype matrix V+ calculates: V+ = (V+V) V(1). Let V is the all learnt samples set vi = 1,…, m. and every column satisfies condition of orthonormality: $v_k^+ v_j = \delta_{ij}$, for all j and k. Where $\delta_{ij}$ is delta of Kronecker. For a sample examination q, parameters of order signify test sampling matching. Class parameter of order for k derives as, $\varepsilon_k = v_k^+ . q$, $k = 1,...,m$.

Due to pseudo inverse over-fitting sometime melting fails to generalize the learning. A penalty function presents as Most Probable Optimum Design (MPOD) to improve the generalization and classify face object pose application (see [30]). Following this modification, the melting combination of similar object patterns into a template is useful for classification. So synergic template is:

$$v_p^+ = E(V^T V + P_1 O + P_2 I)^{-1} V^T \quad (7)$$

I, O, P1, and P2 are identity matrix, unitary matrix, and coefficients of penalty. E is an enhanced identity matrix; every element of E is a row vector of size j as the following:

$$E = \begin{bmatrix} e_1^{n(1)} & e_0^{n(2)} & \cdots & e_0^{n(M)} \\ e_0^{n(1)} & e_1^{n(2)} & \cdots & e_0^{n(M)} \\ \vdots & \vdots & \ddots & e_0^{n(M)} \\ e_0^{n(1)} & e_0^{n(2)} & \cdots & e_1^{n(M)} \end{bmatrix}, \quad e_0^i = (0,...,0), \quad e_1^i = (1,...,1) \quad (8)$$

The kinetic equation based on using q is as follow:

$$\dot{q} = \sum_{k=1}^{M} \lambda_k v_k (v_k^+ q) - B \sum_{k' \neq k} (v_{k'}^+ q)^2 (v_k^+ q) v_k - C(q^+ q)q + F(t)$$

The corresponding kinetic equation for the order parameter

$$\dot{\varepsilon}_k = \lambda_k \varepsilon_k - B \sum_{k' \neq k} \varepsilon_{k'}^2 \varepsilon_k - C \sum_{k'=1}^{M} \varepsilon_{k'}^2 \varepsilon_k$$

Based on the competition, the order parameter which is the strongest will have a victory, that is, to accomplish the pattern recognition purpose. This idea can be realized through a layer-wised network that is depicted in figure 3. Haken suggested the approach with logarithmic mapping-based, FT, and followed coordinates transform technique. The supposed algorithms of learning that assign adjoin vector process of prototype vector. Here, two ways presents regarding assigning prototypes which is utilized synergetic neural networks twice and another one uses key frames of actions for predicting of actions. Attention parameter is also will be determined using quantum particle swarm optimization technique that it will present afterward.

### 3.4 Quantum-Behaved Particle Swarm Optimization for kinetic equation of order parameter

Quantum-behaved particle swarm optimization (QPSO)[39] driven by conceptions from quantum mechanics and particle swarm optimization (PSO), is an algorithm regarding probabilistic optimization adapted from the barebones PSO family[37]. Like PSO by M individuals which each of them is considered as a volumeless particle in an N-dimensional space, by the recent position vector and the velocity vector of particle i, $1 \leq i \leq M$ on the nth iteration represented as $X_{i,n} = (X_{i,n}^1, X_{i,n}^2, ..., X_{i,n}^N)$, and $V_{i,n} = (V_{i,n}^1, V_{i,n}^2, ..., V_{i,n}^N)$ correspondingly. The particle moves based on the behind equations:

$$V_{i,n+1}^j = V_{i,n}^j + c_1 r_{i,n}^j (P_{i,n}^j - X_{i,n}^j) + c_2 R_{i,n}^j (G_{i,n}^j - X_{i,n}^j). \quad (9)$$

$$X_{i,n+1}^j = X_{i,n}^j + V_{i,n+1}^j. \quad (10)$$

For j = 1, 2, . . ., N, where $c_1$ and $c_2$ are known as the acceleration coefficients. The best earlier position vector of particle *i* is shown by $P_{i,n} = (P_{i,n}^1, P_{i,n}^2, ..., P_{i,n}^N)$ (personal best or pbest), and the position vector of the best particle between whole particles in the population is presented by $G_n = (G_n^1, G_n^2, ..., G_n^N)$ (global best or gbest). Following minimization problem will be considered:

$$Min \; f(x), \quad s.t. \quad x \in S \subseteq R^N. \quad (11)$$

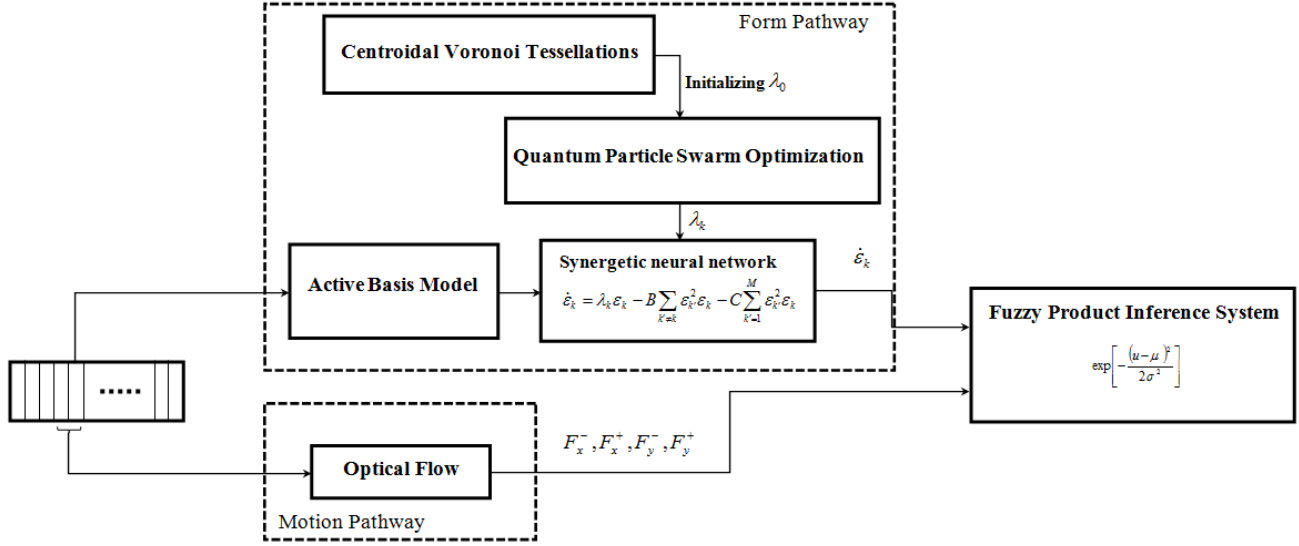

Figure 4. Schematic of recognition in proposed model.

Where Objective function presents by $f(x)$ which in considers as almost continuous function everywhere in feasible space $S$. Thus, for updating $P_{i,n}$ we will have

$$P_{i,n} = \begin{cases} X_{i,n} & f(X_{i,n}) < f(P_{i,n-1}) \\ P_{i,n-1} & otherwise \end{cases}. \quad (12)$$

And $G_n$ can be created by $G_n = P_{g,n}$, where $g = \arg\min_{1 \le i \le M} \{f(P_{i,n})\}$. PSO algorithm may be converging wherever every particle converges to its local attractor $P_{i,n}$ which are defined as:

$$p_{i,n}^j = \varphi_{i,n}^j P_{i,n}^j + (1-\varphi_{i,n}^j) G_n^j, \quad \varphi_{i,n}^j \sim U(0,1) \quad (13)$$

$\varphi_{i,n}^j$ is a sequence of random number between 0, 1, uniformly. Equation shows that stochastic attractor of $i^{th}$ considers in hyper-rectangle and moves by $p_{i,n}$ and $G_n$. Sun et. al. (2012) presented the position of the particle is an updated using equation as follows:

$$X_{i,n}^j = p_{i,n}^j \pm \alpha \left| X_{i,n}^j - \beta \right| \ln(\frac{1}{u_{i,n+1}^j}), \quad (14)$$

Where α is the CE coefficient which is a positive real number and can be adjusted to balance the global and local search of the algorithm within its process. A random numbers uniformly distributed on (0, 1) revealed as sequence is shown by $u_{i,n+1}^j$, varying with $n$ for each $i$ and $j$. Also the mean best (mbest) position is presented by $C_n = (C_n^1, C_n^1, ..., C_n^N)$ which is the average of the best positions (pbest) of all particles, that is,

$$C_n^j = \frac{1}{x} \sum_{i=1}^{M} P_{i,n}^j..$$

### 3.5 Centroidal Voronoi Tessellations for Choosing a Starting the attention parameter

As it is mentioned in the previous section, quantum-behaved particle swarm optimization is applied for finding the optimum order parameter. As it is revealed in the kinetic equation of synergetic neural networks, initialization of the attention parameter ($\lambda_k$) is required to calculate the order parameters updates. Using Voronoi tessellations can be applied as a way to partition a viable space into partitions. The set of generators considers as a group of points in the space which divided into subsets following the approximation of the generators points. Generators are associated with subsets and points are nearer to its corresponding generators rather than any of other generators considering distance function (e.g., the Lz norm). Note that the generators are not very evenly distributed throughout the space. By dividing the spaces into the partitions, several generators set at almost precisely the same point in the space. Although, the centroidal voronoi tessellations by lie the generators at centre of the voronoi cells overcomes to the poor and non-uniform distribution of the some voronoi cells (see [41]).

In this paper, the generators were chosen similar way regarding initialization of initial attention parameters for Particle Swarm Optimization. The proposed approach follows Ju-Du-Gunzburger (JDG) algorithm [44] which produces the feasible computational approximation of

CVTs and its combining the elements of MacQueen's method [43] and Lloyd's method [42]. This algorithm finds the attention parameter initial positions in quantum-behaved particle swarm optimization of order parameter updates more uniformly distributed in the search space.

### 3.6 Combination of two pathways and Max-product fuzzy

The recognition stage schematic regarding classification of human action recognition based on biological inspired model is revealed in Fig. 4. Considering the form features are calculated for both pathways, here main concern is regarding the combination. For that, max product fuzzy method has been utilized transferring the information of both pathways by Gaussian membership function and maximum of their product into fuzzy domain represents the class which action is belonged. Fuzzy logic is a kind of logic having multi-valued that is originated from the theory of fuzzy set found by Zadeh (1965) and it deal with reasoning approximation. It offers high level framework aimed at reasoning approximation which can suitably provide imprecision and uncertainty together in linguistic semantics, model expert heuristics and handles requisite high level organizing principles [50]. Artificial Neural networks refer to computational/mathematical models based on biological neural network and provide self-organizing substrates for presenting information with adaptation capabilities in low level. Fuzzy logic can be a significant complementary method for neural networks because of plausibility and justified for combining the approaches together regarding design classification systems which referred as fuzzy neural network classifier [50], [38]. Also Bourke and Fisher in [52] presented that the max-product gives better outcomes than the usual max-min operator. Consequently, similar algorithms by having effective learning scheme have been mentioned by others
[51], [53], [54] using the max-product composition later. Here in this paper, fuzzy Max-production composition is applied inside the synergetic neural network regarding form and motion pathways aggregation. It means the initial order parameter will be obtained by combination of these two pathways for better decision making.

**Definition of motion pathway classes in different action** all possible action of human object optical flow captured and store in a database considered as references. Each references optical flow data in every action assign in a specific amount of optical flow regarding specific actions which will be assigned by interpretation of an operator (human observer) as a training map, generally description of which could be called Operator perceived activity (OPA) [55]. Considering that mean and standard deviation of every class are different from each other, operator comments on each of reference data will be different and classification among the classes will be done.

**Max-product fuzzy classifier** Fuzzy production among two pathways classification is carried out through general strategy of having result estimated as following composition from both pathways presented as below:

$\mu_{FP\omega}(\dot{\varepsilon}_k, C_i, t)$ and $G_{MP\omega}(f_k, C_i, t)$ are outputs of quaternion correlator in enrolment stage belong to form and motion pathways, respectively. Fuzzification is done through Gaussian membership function as activation functions:

$$G_{MP\omega}(f_k, C_i, t) = \exp\left[\frac{FP\omega(\dot{\varepsilon}_k, C_i, t) - \mu_{FP\omega}}{\sigma^2}\right] \quad (15)$$

Where $\dot{\varepsilon}_k$ comes from unbalanced order parameter $k^{th}$ subject in frame time $t$ belongs to $c_i$ estimate from active basis model as form pathway and directly relates to $\lambda_k$ as its $k^{th}$ attention parameter tuned by quantum-behaved particle swarm optimization in the training stage. Also for motion pathway membership is Gaussian functions deviation as below:

$$G_{MP\omega}(f^{\pm}_k, C_i, t) = \exp\left[\frac{(MP\omega(f^{\pm}_k, C_i, t) - \mu_{MP\omega})^2}{\sigma^2}\right] \quad (16)$$

$$G_{MP\omega}(f^{\pm}_k, C_i, t) = G_{MP\omega}(f^{-}_k, C_i, t) \times G_{MP\omega}(f^{+}_k, C_i, t)$$

$$G_{MP\omega}(f^{\pm}_k, C_i, t) = G_{MP\omega}(f^{-}_{y,k}, C_i, t) \times G_{MP\omega}(f^{+}_{y,k}, C_i, t)$$

$$\mu_{MP\omega}(f^{\pm}_{\tau,k}, C_i, t) = G_{MP\omega}(f^{\pm}_{x,k}, C_i, t) \times G_{MP\omega}(f^{\pm}_{y,k}, C_i, t)$$

Where $f^{\pm}_{\tau,k}$ is positive or negative (direction) flow in $\tau = x_{or} y$ of $k^{th}$ subject in frame time **t** as representation of motion pathway amount for every class $c_i$. $\mu_{FP\omega}(\dot{\varepsilon}_k)$ and $\mu_{MP\omega}$ are mean value and is standard deviation of both pathway.

(2) Determine the value of product by considering trained attention parameter in form pathway and trained parameters of motion pathway in $k^{th}$ subject in frame time **t**.

$$\mu_{M\omega'_k} = G_{FP\omega}(\dot{\varepsilon}_k, C_i, t) \times \mu_{MP\omega}(f^{\pm}_{\tau,k}, C_i, t) \quad (17)$$

(3) Gather the values of product in an array similar for amount of membership in class of every action with both pathways separately:

$$x_{\mu''} = \begin{bmatrix} \mu_{P\omega'_1,C_1} & \mu_{P\omega'_1,C_2} & \cdots & \mu_{P\omega'_1,C_i} \\ \mu_{P\omega'_2,C_1} & \mu_{P\omega'_2,C_2} & \cdots & \mu_{P\omega'_2,C_i} \\ \vdots & \vdots & \ddots & \vdots \\ \mu_{P\omega'_k,C_1} & \mu_{P\omega'_k,C_2} & \cdots & \mu_{P\omega'_k,C_i} \end{bmatrix} \quad (18)$$

(4) Presents output array and a set of produced membership amounts reveals the belonging degrees to every class $C_i$. The biggest amount represents the degree of belong to each classes and winner take all.

(5) Determine which element in classification matrix $Y_\mu$ has maximum degree of the membership among all $i$ classes.

$\psi$ = number of element position in classification matrix $Y_\mu$ which has the maximum value with $C_i$ class. $\psi$ presents the assigned number of reference image in database.

(6) Following one fuzzy IF-THEN rule, perform defuzzification:

$\mathbf{R^1}_s$: IF $\mu_{P\omega'_\alpha,C_i}$ from subject $\alpha$ in class has maximum degree in membership function as compare with others, THEN subject classified as class $C_i$.

4. **EVALUATION AND RESULTS**

Experimental results are extensively presented to reveal the effectiveness and estimate the ability of proposed model to human action recognition task.

**4.1 Biological Inspire Model and Relation to existing methods**

Co-operation among information attained from two processing streams occurs at few levels in the mammalian brains [15], [16] and it can simplifies the aggregation of model (for instance in STS level [17]) and improve recognition performance. Holonomical features considering both pathways for predefined action templates. In form pathway, proposed approach followed Karl Pribram's holonomic theory which is based on evidence that dendritic receptive fields in sensory cortexes are described mathematically by Gabor functions [49] that is vastly utilized by active basis model[5]. As it is aforementioned, primary stage is includes local (in V1 cell) and model detectors (Gabor like filters) in sixteen (including eight preferred) orientations and by proper scale depend on receptive field (see [10], [19]). Active basis model also played the role of snapshots detectors regarding human body shapes model finding like with area IT (inferotemporal cortex) of the monkey where view-tuned neurons located and model of complex shapes tune [22] which is implemented applying synergetic neural network. Especially unbalanced synergetic neural network by tuning optimized attention parameters works as view-tuned neurons in area IT and Snapshot neurons regarding providing independency in scale and position.

Proposed model follows getting the modelling and adjusted through training as key frames. Utilizing optical flow outcome and infer it with information obtained from form pathway, presented approach covered high level integration of snapshot neurons outcomes with motion pattern neurons information. Furthermore, active basis model used computational mechanism regarding recognized human object form which is follows up the neurobiological model in dorsal stream located in visual cortex(V1)[3],[7],[8]. As local direction has been organized in initial level of form pathway and Gabor like modelling detector methods i.e. active basis model have good constancy by modelling cells in mentioned part [18]. Sixteen directions and two spatial scales by two differentiators and finding information of local direction in the pathway and complex-like cells having independent form features which are appropriate for form pathway will be done by using mechanism of proposed neurophysiological plausible model.

In motion pathway, biological movement has consistency with neurophysiological information of neural detectors in MT and V1 regarding motion and direction that is done by applying optical flow [3]. Estimation of local motion is also directly computed from optical ow which is response of motion selective neurons in the area of MT. In areas of MT, MSTd, MSTl and some parts of dorsal steams and probably in kinetic occipital area (KO) motion selects by opposite directions [3] which are modelled by $F_x^-$, $F_y^-$, $F_x^+$, and $F_y^+$. Considering maximum pooling motion and its amount of using fuzzy Gaussian membership function for each directions of optical flow and fuzzy product decision membership can be a very good presenter for form pathway and third level motion pathway by snapshot neurons and applied approach is good combination of two pathways for vertical stream in V2, V4 projection and primary visual cortex(V1) which has been lowed here.

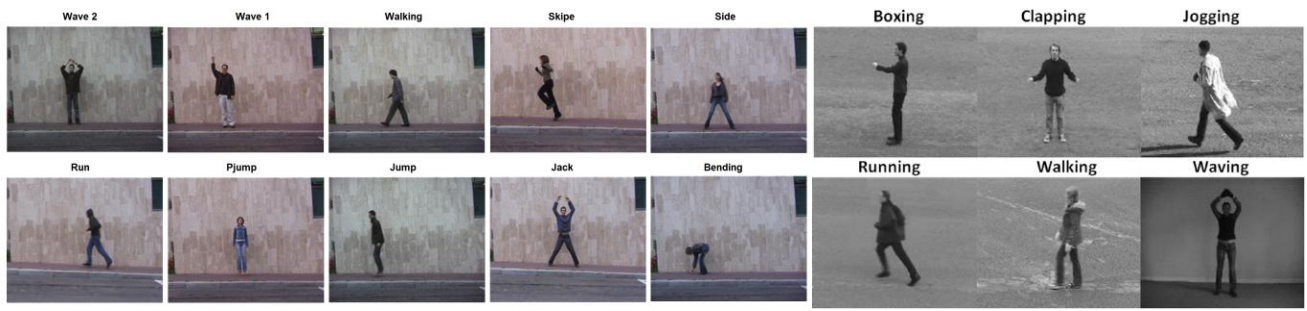

Figure 5. Schematic of recognition in proposed model.

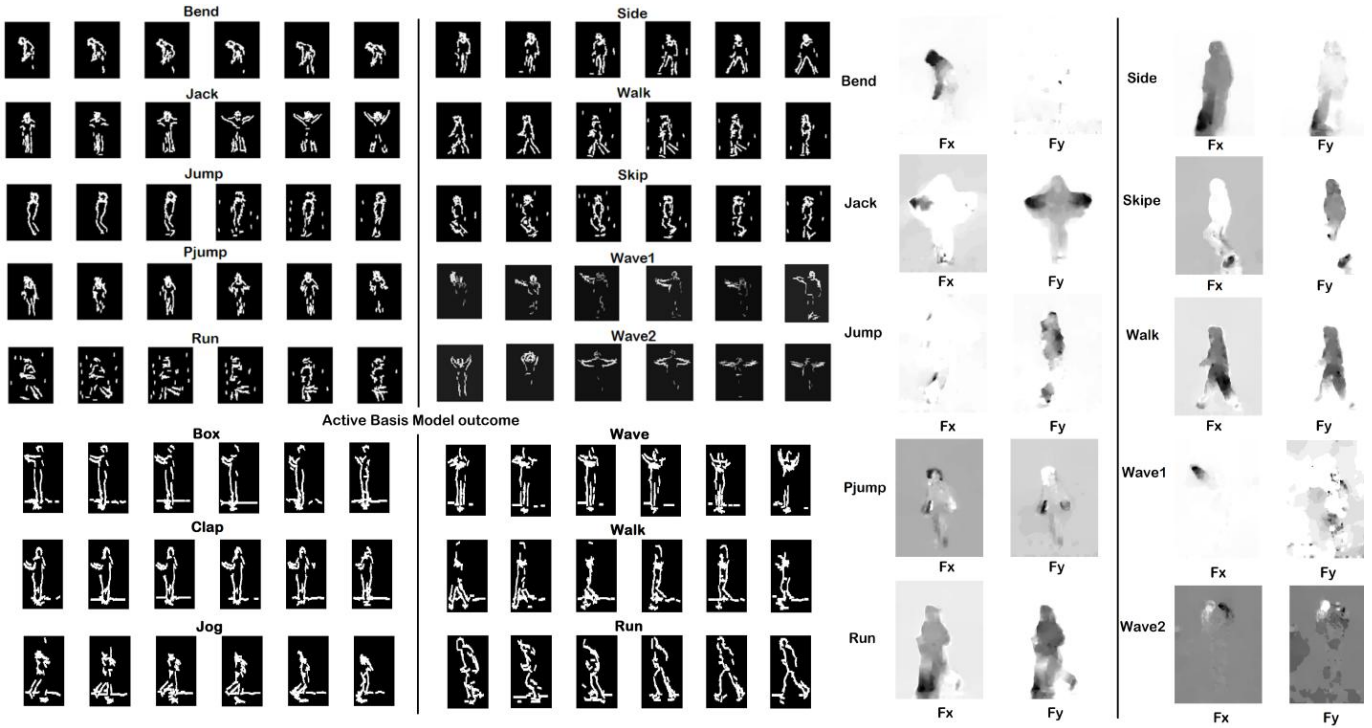

Figure 6: Schematic of recognition in proposed model.

The proposed model is like current techniques follows hierarchical feed-forward designs like [7] and specially tries to develop a model that it follows neurobiological motion processing in visual cortex and basically follows [3]. Object recognition task in form pathway has been changed within the researchers work from spatiotemporal features in [7] and original Gabor filter [8] to the model by using active basis model. However, active basis model has basic characteristic of previous features and basically uses Gabor wavelet but it decreases matching operation. It activates on the limited clutters and ensures the important amounts in points of interest which falls on person subject. In aspects of used features, layer-wised optical flow [4] which is simply silhouette form regarding motion and form of subject and better combination of two pathways using fuzzy inference theory and classifying by synergetic neural network tuned by quantum particle swarm optimization that it makes the model more biological.

**4.2 Data Sets**

KTH action dataset [31] as the largest human action dataset including 598 action sequences that it comprises six types of single person actions as boxing, clapping, jogging, running, walking, waving. These actions perform by 25 people in different conditions: outdoors (s1), outdoors with scale variation (s2), outdoors with different clothes (s3), and indoors with lighting variation (s4). Here, using down-sampling the sequences resolutions become to 200 142 pixels. For our approach, we used 5 random cases (subjects) for training and making form and motion predefined templates. As it is mentioned in literature, KTH is a robust intra-subject variation with large set whereas camera for taking video during preparation had some shacking and it makes the work with this database very difficult. Moreover, it has four scenarios which are independent, separately trained and tested (i.e., four visually different databases, which share same classes). Both alternatives have been run. For considering symmetry problem of human actions, there is a mirror function for sequences along with vertical axis which can be available for testing and training sets. Here all possible overlapping of human actions within training and testing sets has been considered (e.g. one video has32 and 24 action frames.)

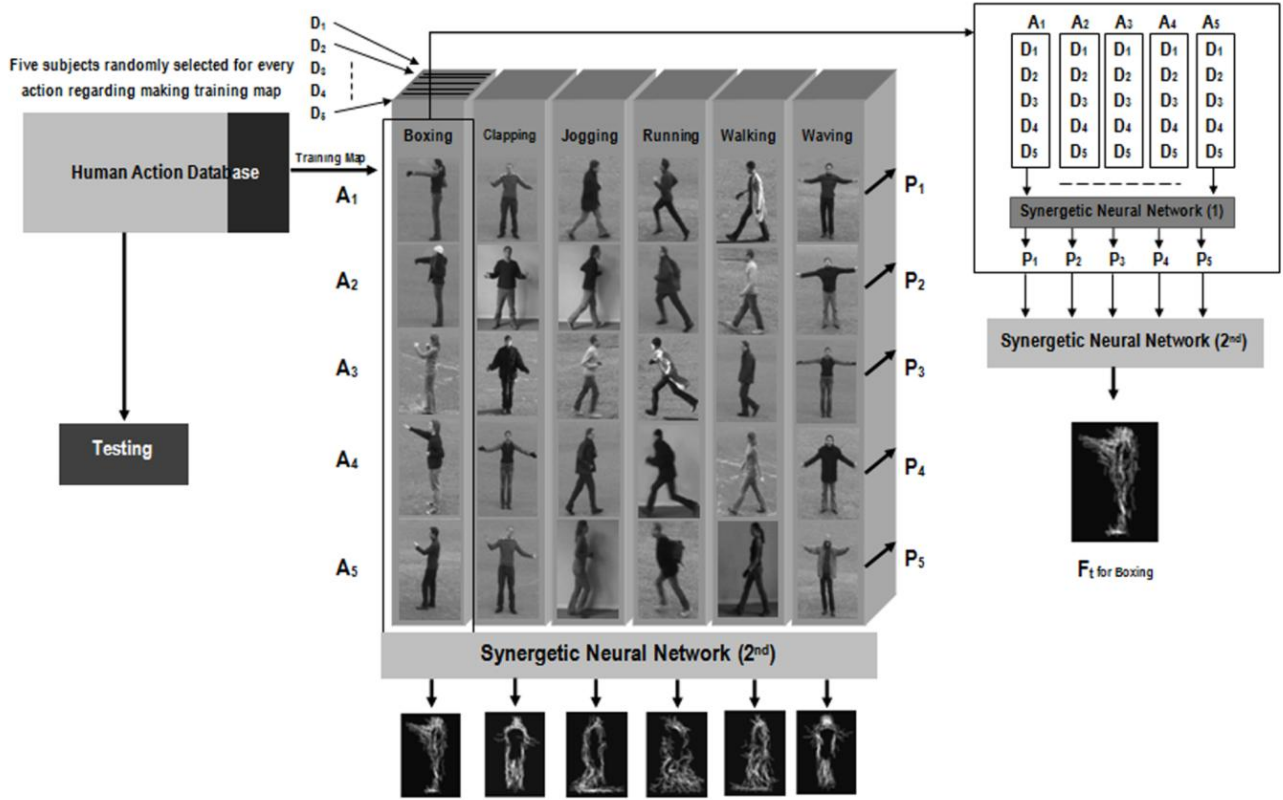

Figure 7: Figure shows the procedure of making the action active basis templates by applying two times synergetic neural networks on the training map which calculates from randomly selected video frames from KTH human action database.

Weizmann human action database [40] comprises nine types of single person actions, and having 83 video streams reveals performing nine different actions: running, galloping sideways, jumping in place on two legs, walking, jack, jumping forward on two legs, waving one hand, waving two hands, and bending. We track and stabilize figures using background subtraction masks that come with this data set. Sample frames of this data set is shown in Fig. 5.

The above mentioned two data sets have been widely utilized to estimate methods/techniques designed for action recognition. Though, they only concentrate on single person actions recognition e.g., clapping, walking. To understand advantages of proposed approach. On our testing data sets, we illustrate experimental results on using synergetic neural network in balanced and unbalance modes and a comparison among previous work that proposed biological human action models. There was a comparison for classification between balanced and unbalanced classifying of Form pathway along with consideration of accuracy for form and motion pathways after application of fuzzy product between these two pathways.

Proposed model is efficient and computational cost will be due to feature extraction regarding two pathways form and motion features applying active basis model and optical flow respectively.

After optimization part regarding tune of attention parameter in unbalanced synergetic neural network for form pathway, system infers on a new video only takes a few seconds in our unoptimized MATLAB implementation which it is combined
by existing codes for motion and form pathway in MATLAB/C [56], [57]. Subsequently by computing features in both pathways with the different setting which is aforementioned, system trained and tested correspondingly. For a specified test sequence, the action label is assigned to the action frames. Then the accuracy of classification is specified by

## KTH Confusion Matrix (Figure 8, top-left)

|         | Boxing | Clapping | Jogging | Running | Walking | Waving |
|---------|--------|----------|---------|---------|---------|--------|
| Boxing  | 0.95   | 0.02     | 0       | 0       | 0       | 0.03   |
| Clapping| 0.16   | 0.69     | 0       | 0       | 0       | 0.15   |
| Jogging | 0      | 0        | 0.66    | 0.20    | 0.14    | 0      |
| Running | 0      | 0        | 0.10    | 0.88    | 0.02    | 0      |
| Walking | 0      | 0        | 0.26    | 0.05    | 0.70    | 0      |
| Waving  | 0.21   | 0.04     | 0       | 0       | 0       | 0.74   |

## KTH Confusion Matrix (Figure 9, top-right)

|         | Boxing | Clapping | Jogging | Running | Walking | Waving |
|---------|--------|----------|---------|---------|---------|--------|
| Boxing  | 0.89   | 0        | 0       | 0       | 0       | 0.11   |
| Clapping| 0.05   | 0.83     | 0       | 0       | 0       | 0.11   |
| Jogging | 0      | 0        | 0.79    | 0.08    | 0.12    | 0      |
| Running | 0      | 0        | 0.05    | 0.95    | 0       | 0      |
| Walking | 0      | 0        | 0.17    | 0       | 0.82    | 0      |
| Waving  | 0.14   | 0.14     | 0       | 0       | 0       | 0.72   |

## Weizmann Confusion Matrix (Figure 8, bottom-left)

|       | Bend | Jack | Jump | Pjump | Run  | Side | Skip | Walk | Wave1 | Wave2 |
|-------|------|------|------|-------|------|------|------|------|-------|-------|
| Bend  | 0.8  | 0    | 0    | 0     | 0    | 0    | 0    | 0    | 0.2   | 0     |
| Jack  | 0    | 0.8  | 0    | 0     | 0    | 0    | 0    | 0    | 0.1   | 0.1   |
| Jump  | 0    | 0.2  | 0.6  | 0.1   | 0    | 0    | 0    | 0    | 0.1   | 0     |
| Pjump | 0    | 0.1  | 0    | 0.9   | 0    | 0    | 0    | 0    | 0     | 0     |
| Run   | 0    | 0.24 | 0    | 0.12  | 0.50 | 0    | 0    | 0.12 | 0     | 0     |
| Side  | 0    | 0    | 0    | 0     | 0    | 0.62 | 0.25 | 0    | 0.12  | 0     |
| Skip  | 0    | 0    | 0.12 | 0.12  | 0.12 | 0    | 0.50 | 0    | 0.12  | 0     |
| Walk  | 0    | 0.22 | 0    | 0     | 0    | 0    | 0.11 | 0.56 | 0     | 0     |
| Wave1 | 0    | 0    | 0    | 0     | 0    | 0    | 0    | 0    | 0.67  | 0.33  |
| Wave2 | 0    | 0    | 0    | 0     | 0    | 0    | 0    | 0    | 0     | 1.00  |

## Weizmann Confusion Matrix (Figure 9, bottom-right)

|       | Bend | Jack | Jump | Pjump | Run  | Side | Skip | Walk | Wave1 | Wave2 |
|-------|------|------|------|-------|------|------|------|------|-------|-------|
| Bend  | 1.00 | 0    | 0    | 0     | 0    | 0    | 0    | 0    | 0     | 0     |
| Jack  | 0    | 0.89 | 0    | 0     | 0    | 0    | 0    | 0.11 | 0     | 0     |
| Jump  | 0.11 | 0    | 0.89 | 0     | 0    | 0    | 0    | 0    | 0     | 0     |
| Pjump | 0    | 0    | 0    | 1.00  | 0    | 0    | 0    | 0    | 0     | 0     |
| Run   | 0    | 0    | 0    | 0     | 0.75 | 0    | 0.12 | 0.12 | 0     | 0     |
| Side  | 0.13 | 0    | 0    | 0     | 0    | 0.75 | 0    | 0    | 0.12  | 0     |
| Skip  | 0.23 | 0.11 | 0    | 0     | 0    | 0    | 0.56 | 0    | 0     | 0     |
| Walk  | 0    | 0    | 0.23 | 0     | 0    | 0    | 0.11 | 0.56 | 0     | 0     |
| Wave1 | 0    | 0    | 0    | 0     | 0    | 0    | 0    | 0    | 1.00  | 0     |
| Wave2 | 0    | 0    | 0    | 0     | 0    | 0    | 0    | 0.33 | 0     | 0.67  |

Figure 8: Confusion matrices representing the accuracy of recognition in KTH and Weizmann data set, using multi-prototype human action templates.

Figure 9: Confusion matrices for recognition of human action in KTH and Weizmann data sets applying second scenario.

$$Acc = \frac{Numbers\ of\ correct\ classified\ frames}{Total\ number\ of\ frames} \quad (19)$$

The algorithm correctly classifies the most of actions (see the confusion matrices revealed below). Most of the occurred mistakes are in recognition of running, jogging also boxing, clapping and waving. The intuitive reasoning for it is because of similarity among these two groups of action. On testing our data bases, the confusion matrices have been obtained for two proposed scenario regarding application of our methods and overall accuracy for both per-fame and per-video classification. Confusion matrices (per-video or perframe) are both proposed scenarios have similar patterns, so we only reveals one confusion matrix for every dataset. Result of each scenario has been mentioned in table 1 which represents accuracy of proposed techniques as compare with some previous methods on same data set. But this comparison is not precise because of differences in experimental setups as presented results are comparable with state-of-the-art techniques whereas considering the various methods has all sorts of the differences in their setups like unsupervision or supervision, with tracking (similar with [64,65,67]) or without it, subtraction of the background, or considering multiple actions recognition, etc.

In term of biology, movement contains corticofugal pathways from both peristriate cortex (V2) and striate cortex (V1). The peristriate (V2) and striate (V1) cortices are mutually linked and there are only minor, though important, differences in their receptive properties. In theory of holonomic brain, peristrate (V2) and straite (V1) are narrowly coupled collaborating system through virtue of both reciprocal cortico-connectivity plus connection of them to brain stem tectal region. It is upon this carefully joined organism that extra compound perceptual procedure converges. Convergence locus is region of brain stem tectal close to colliculi that it is in turn connects to colliculi. About vision, superior colliculus connections to neurons in striate cortex (V1) that it shows complex receptive fields, complete the circuit [58]. Also, a set of receptive fields is particularly sensitive to processing movement in the visual input, specifically virtual movement of one portion of input through respect to another. This sensitivity to relative movement is critical to the formation of object-centered spaces. Another set is principally sensitive to comparative movement among somatosensory and visual inputs. Receptive fields of these neurons are straightly comprised in the egocentric action spaces formation (See [49]). Considering aforementioned in term of biological, the proposed model has been considered two structures for V1 information of form pathways for finding the shape and form of human objects by input of original frame after active basis model application and at the end comparison has been done among each of two configurations.

In V2, proposed method used local representation and action sequence is selected by its location. The response of active basis model is directly used for classification of action.

### 4.3 Multi-prototype human action Templates

In this scenario, recognition of human action pattern in form pathway has done by one predefined template which attained by applying synergetic neural network prototypes. First, we performed multi-prototype predefined templates for each human action obtained applying synergetic neural network on human action image. For making training map of every action, we divide every human action sequence to five primitive basic movements. Once can create whole action sequence using these five basic actions. Besides, considering the style invariance difficulties regarding diverse object in same action, proposed training map attains using five different subjects from targeted human action databases. For easing the explanation, we consider five snippets in different actions A1-A5 and each subject from targeted database D1 - D5. First, synergetic neural network applies to A1 in D1 - D5 and outcome shows by P1as first prototype obtains from first action snippet. Number of prototypes will be completed by applying synergetic neural network and calculating the residual prototypes that they have called P1 - P5. Calculated prototype images considering style invariance represent one action within five snapshots. Afterward, these prototypes melt together using second time synergetic neural network for attaining the final prototypes which each of them represents the specific action within different action snippets and considering style invariance property. Let Ft represents outcome of melting P1 - P5 in specified action. The final prototype images for each human action and the application of synergetic neural network procedure to make training map is presented in Fig. 7. Recognition result of first scenario is revealed in fig. 8. Consider that there are two categories using dissimilar paradigms, which cannot be straightly compared. Here, experimental result of the proposed approach is presented. As KTH and Weizmann human action database have been used for benchmarking the accuracy of consistency with the set of experiments used in [7],[8], [33], [34], [35] , [36], we made set of our training map and test of proposed technique on entire data set, in which mixture of four scenarios videos were together(for KTH data set). The data set split into a set of training-map with five randomly selected subjects and a test part by residual subjects. Afterward, we measured the performance average over five random splits following their frames numbers.

### 4.4 Second Scenario for applying action templates

Our biological motivated model in form pathway is very much inspired of computer visional bag-of-words method regarding problem if object recognition. Regular concept of mentioned approaches comprises of extracting the features in specific local from a set of image frames for every action, assembling a codebook of visual action words with vector quantization, and construction of model regarding representation of action by utilizing four key frames of each action. Although these models are not certainly correct a these are consider a set of patches which are locally selected and may ignore many structure, they have been acknowledge as efficient object recognition methods [59], [60]. In proposed approach, we utilized some frames as key frames (words) for recognition of action in whole action frames for recognizing human actions. In proposed approach, every frame of action video is consigned as one visual word by considering similarity of each with action codebook. Like problems of object recognition, particular structures have been missed by moving to this representation [36]. This method has a good performance while the local distribution of action sequence is very similar to targeted action and very different from other sample sequences which are in same action frames but different categories. Concisely, variance of intraclass is big and variance of interclass is less. Especially in case of single person human action recognition, the variance of intraclass is smaller than the multiperson [33]; therefore, its application has been performed significant in proposed approach.

| Methods | Accuracy(percent) | Year |
|---|---|---|
| Schuldt. [34] | 71.72 | 2004 |
| Niebles. [35] | 83.33 | 2006 |
| Jhuang. [7] | 91.7 | 2007 |
| Schindler. [8] | 92.7 | 2008 |
| Wang. [36] | 91.2 | 2009 |
| Zhang. [33] | U-SFA:84.67<br>S-SFA:88.83<br>D-SFA:91.17<br>SD-SFA:93.50 | 2012 |
| Proposed Model | $1^{st}$ scenario:78.05<br>$2^{nd}$ scenario:83.34 | 2013 |

Table 1: The results of recognition by proposed method has presented along with comparison among previous methods on the KTH human action dataset.

| Methods | Accuracy(percent) | Year |
|---|---|---|
| Schuldt. [34] | 72.8 | 2004 |
| Niebles. [35] | 72.8 | 2006 |
| Jhuang. [7] | 98.8 | 2007 |
| Schindler. [8] | 100 | 2008 |
| Wang. [36] | 100 | 2009 |
| Zhang. [33] | U-SFA:86.67<br>S-SFA:86.40<br>D-SFA:89.33<br>SD-SFA:93.87 | 2012 |
| Proposed Model | $1^{st}$ scenario:70<br>$2^{nd}$ scenario:81.03 | 2013 |

Table 2: Comparison of the proposed approach and previous methods for Weizmann human action dataset.

**4.5 Evaluation of Quantum-Behaved**

Particle Swarm Optimization Results have previously mentioned are in balanced mode of synergetic neural network that has been done for better comparison among both scenarios of form pathway whereas using quantum particle swarm optimization has very good tuning performance for attention parameters. Attention parameter in balance mode is constant and equal to one. While a procedure is working to find solution of the problem at hand, one of the most significant issues is how to choice its parameters and initiating them. For initial attention parameter, Centroidal Voronoi Tessellations has been used. The value of _ is constant and static. The algorithm has run for 500 echoes for 20 particles as population size and 20 times.

**4.6 Evaluation**

After converging the algorithm attention parameters have been used in unbalanced synergetic neural network for results of form pathway. Evaluation of proposed approach through two human action data sets has done and confusion matrices are previously shown. Here, we show the performance of proposed method with compare our results with previous approaches on same data set as revealed in Table 1 and Table 2. Proposed method performances on KTH and Weizmann data sets are saturating state-of-the-art methods reach good and comparable results. The comparison of our method as biological inspired model with state-of-the-art (with or without biological point of view) listed in the Table 1 and Table 2.

Also, we should note that different methods listed in Table 1 have all sorts of variations in their experimental setups, e.g., different splits of training/testing data, whether some preprocessing (e.g., tracking, background subtraction) is needed, with or without supervision, whether per-frame classification can be done, whether a method handles multiple action classes in a video, etc. Results of our methods are comparable specially in term of robustness to other state-of-the-art approaches, although we accept the fact that comparing with some of methods are not absolutely fair, meanwhile their method does not completely covered biological point of view (e.g. [34]). But considering [33] as a technique which is biologically inspired, is revealed that proposed model is very near accuracy.

**4.7 Discussion**

The correctly classified sequences are reported as the highest results in literature. To place proposed technique in this context, we have presents it with state-of-the-art. Our method similar with other method which is frame-based run for all frames of action sequences. Individual labels obtained from training map simply compare to a sequence label through majority voting (it is like a bag-of-frames model and like [8]). Comparison with state-of-the-art has been done and it reveals in table 1, and table 2. It is noticeable that original frames are adopted as input of system and using different frames can have less performance by considering random location of Gabor beams on human object in different frames. Training map dataset (multi-prototype template set) was comprised five frames of video snippets randomly obtained from the mixture dataset for the case of multi-templates experiment. Also in second scenario, four key-frames have been precisely selected from videos which randomly selected for every action. Fig. 8 and Fig. 9 present the classification confusion matrices for the KTH and Weizmann data sets. The row of confusion matrices represent the corresponding classification results, while each column signifies the instances to be classified. In terms of contribution, we can mention applying active basis model in form path way as first time is utilized in biological model and fuzzy inference system regarding combination of two pathways made proposed model novel.

However, the natural question (see [8]) regarding whether this combination is necessary and how, is still there and still trying to improve the model and make it more accurate. We have performed experiments for presented model, in which we have modified form pathway and made it combined with motion features and completed a relation for these two almost independent feature pathways together which revealed promising results. Proposed approach is robust which has a major strength comparing with other human action recognition methods which used similar biological model. Plus It combined form and motion pathways with respect to original model. Regarding combination, a question may arise that is it necessary to combine these two pathways? And which combination form is better?. By fuzzy inference system the information attained from motion pathway helps form pathway or other way around. However, combination of motion and form, regularly overtakes both motion and form separately, in most of the experiments conducted, combining the information of these two pathways takes place in final decision part(see [3], [7], [8]).

Besides, relative feed-forward structure from input data-stream till final decision does not change and is similar across different data-sets among two independent sets of features compute (see figure1 in [3] and _gure2 in [8]). Here, we have presented that by considering original model topology regarding both pathways; extracted features for each pathway can be relatively utilized in other pathway and configuration of both pathways has been modified by using the fuzzy inference technique.

5. CONCLUSION

In this paper, biological inspired model based on inter-relevant calculated motion and form features tested by applying for human action recognition task has presented. Principally, we have defined form features applying active basis model as form extractor in form pathway and optical flow as flow detector in motion pathway for video sequence stream. Unbalanced synergetic neural networks has been utilized for classification of shapes and structures of human objects along with tuning of quantum particle swarm optimization by initiation of Centroidal Voronoi Tessellations utilized and justified as a good tools in form pathway. At the end, decision has been done by combination of final outcomes of both pathways in fuzzy inference domain and fusion of these two brain pathways considering each feature sets to Gaussian membership functions and then fuzzy product inference. Two configurations have been proposed for form pathway: applying multi-prototype human action templates using two time synergetic neural network for obtaining uniform template, and second scenario that used a model motivated from bag-of-words and abstracting the human action in four key-frames. Experimental result of proposed model has shown promising accuracy and robust performance has been shown using KTH, Weizmann data sets. Furthermore, It has good performance on different datasets and its training done by less computational load regarding final prototype template learning. However, initialization of attention parameters needs more time to find the proper attention parameters. As open-questions, that it continues and should be scrutinized is, how can we diminish the computational load for training of the model?, Is it need to improvement? Future work will extend proposed approach better integration of present form and motion information in two pathways. Another extension is to find more accurate way regarding classifier.

ACKNOWLEDGEMENT

The authors would like to thank Ce Liu for providing code for layer-wised optical flow [4] as well as Ying Nian Wu for active basis model code [5].

This research was sponsored by grants from: contract No. UM.C/HIR/MOHE/FCSIT/10, High Impact Research (HIR) foundation in University Malaya (UM) Malaysia.